\documentclass{article}
\usepackage{amssymb}
\usepackage{amsmath}
\usepackage{amsfonts}
\usepackage{graphicx}
\usepackage{subcaption}
\usepackage{mathtools}
\usepackage{booktabs}
\usepackage[table]{xcolor}
\usepackage{bm}    
\usepackage{makecell}
\usepackage{multirow}
\usepackage{authblk}
\usepackage{url}
\usepackage[margin=1.2in]{geometry}

\newcommand{\R}{\mathbb{R}}

\begin{document}
\title{Fast Multipole Attention: A Scalable Multilevel Attention Mechanism for Text and Images}
\author[1]{Yanming Kang}
\author[1]{Giang Tran}
\author[1]{Hans De Sterck}
\affil[1]{University of Waterloo}
\date{September 3, 2025}
\maketitle

\begin{abstract}
While Transformer networks benefit from a global receptive field, their quadratic cost relative to sequence length restricts their application to long sequences and high-resolution inputs. We introduce Fast Multipole Attention (FMA), a divide-and-conquer mechanism for self-attention inspired by the Fast Multipole Method from n-body physics. FMA reduces the time and memory complexity of self-attention from $\mathcal{O}\left(n^2\right)$ to $\mathcal{O}(n \log n)$ and $\mathcal{O}(n)$ while preserving full-context interactions.

FMA contains a learned hierarchy with $\mathcal{O}(\log n)$ levels of resolution. In this hierarchy, nearby tokens interact at full resolution, while distant tokens engage through progressively coarser, learned basis functions. We have developed both 1D and 2D implementations of FMA for language and vision tasks, respectively. On autoregressive and bidirectional language modeling benchmarks, the 1D variant either matches or outperforms leading efficient attention baselines with substantially lower memory use. With linear complexity, the 2D variant demonstrates superior performance over strong vision transformer baselines in classification and semantic segmentation tasks.

Our results confirm that the multilevel attention implemented by FMA allows Transformer-based models to scale to much longer sequences and higher-resolution inputs without loss in accuracy. This provides a principled, physics-inspired approach for developing scalable neural networks suitable for language, vision, and multimodal tasks. Our code will be available at \url{https://github.com/epoch98/FMA}.

\end{abstract}

\section{Introduction}
The Transformer deep learning architecture \cite{attention} was first introduced in the context of neural machine translation \cite{Bahdanau} and has since been widely adopted across various domains, including image recognition \cite{vit}, music generation \cite{musictransformer}, speech recognition \cite{conformer}, protein structure prediction \cite{alphafold}, and as the backbone of foundation models \cite{gpt3,radford2021clip,reed2022generalist}.

In computer vision, the success of the Vision Transformer (ViT) \cite{vit} sparked a family of 2D transformers such as Pyramid Vision Transformer (PVT) \cite{pvt}, Swin Transformer \cite{swin}, and SegFormer \cite{segformer}.
These hierarchical or pyramid designs adapt self-attention to high-resolution images, achieving state-of-the-art accuracy in classification, detection, and semantic-segmentation benchmarks while controlling computational cost.
Our proposed Fast Multipole Attention (FMA) extends this line of work, offering a principled and dimension-independent approach inspired by the Fast Multipole Method from n-body physics. 
This method achieves linear complexity for both 1D and 2D inputs, and enables efficient long-context processing in language and high-resolution vision tasks.
% \subsection{Self-attention Mechanism}
 
The core of the Transformer architecture is the \textit{self-attention mechanism}, which is briefly described as follows.
Consider an input sequence of tokens with length $n$, for example, a sentence with $n$ words. Assume the sequence is represented by a matrix $X\in \mathbb{R}^{n\times {d}}$, where row $i$ of $X$ represents token $i$ embedded in a vector space with dimension $d$.
The self-attention mechanism learns, for each token $i$, a query vector, a key vector and a value vector, each embedded in a vector space with dimension $\hat{d}$. In the neural network, the matrices of queries $Q\in\mathbb{R}^{n\times \hat{d}}$, keys $K\in\mathbb{R}^{n\times \hat{d}}$ and values $V\in\mathbb{R}^{n\times \hat{d}}$ are obtained from the input sequence embedding $X$ using learned linear projection matrices $W_Q,W_K,W_V\in \mathbb{R}^{d\times \hat{d}}: Q=XW_Q$, $K=XW_K$, and $V=XW_V$.

The \textit{(unnormalized) attention matrix} computes similarity scores between queries and keys:
\begin{align}
            C  = QK^\top, \label{C}
\end{align}
where matrix element $c_{ij}$ encodes the similarity between the query vector of token $i$ and the key vector of token $j$. The \textit{scaled dot-product attention} is defined by
    \begin{align}
        \hat{X}=\mathtt{ATTN}(Q,K,V)={\mathtt{softmax}}{\left(\frac{C}{\sqrt{\hat{d} }}\right)}V.  \label{attn}
    \end{align}
Here {$\mathtt{softmax}$} is the standard normalized exponential function that is applied to each matrix row.
Essentially, for each token $i$ the attention mechanism
calculates a convex combination of the value vectors in the rows of $V$ with coefficients obtained
from inner products between query and key vectors.
In this way, self-attention converts the input sequence $X\in \mathbb{R}^{n\times {d}}$ into a transformed sequence $\hat{X}\in \mathbb{R}^{n\times \hat{d}}$, where each new token is a weighted linear combination of the value vectors of all input tokens. The weight matrices $W_Q,W_K,W_V$ are trained to minimize the loss function of the Transformer.

Unlike recurrent neural network layers and convolutional layers, self-attention provides the Transformer architecture with a global receptive field. However, this advantage comes with quadratic time and space complexity: calculating the attention matrix according to Eq.\ (\ref{C}) requires $\mathcal{O}(n^2 \hat{d})$
time and $\mathcal{O}(n^2)$ space. This becomes a major computational bottleneck for large $n$ and limits the maximum context size feasible in practice.

To address this problem, many methods focusing on improving the efficiency of self-attention have been proposed.
One line of work is sparsification of the attention matrix with either fixed patterns \cite{sparse, longformer} or clustering methods \cite{reformer, Routing}.
However, the ability of attention to capture information from the entire sequence is impaired by these modifications. 
Another direction is to linearize attention
either by kernelizing \cite{performer, linear}
or replacing $\mathtt{softmax}$ with a linear operation \cite{zhen2022cosformer}, to avoid  forming the
attention matrix explicitly.
Although these approaches have complexity
linear in $n$, their main disadvantage is that
in autoregressive models they require recurrent rather than parallel training along the sequence length dimension, which limits hardware utilization, making them difficult to scale to very long sequences \cite{tay2022efficient}.

In this work, we propose an efficient  variant of attention called \textit{Fast Multipole Attention (FMA)}, which
uses different resolutions according to the distance
between the input and output tokens. 
This multilevel strategy is inspired by fast summation methods from $n$-body physics and the Fast Multipole Method \cite{Barnes1986-ah,greengard1987fast, beatson1997short,Martinsson2015}.
Specifically,
 in the neighborhood of the query, attention is
 calculated with keys and values in full resolution.
 Moving away from the query, keys and values are 
 grouped and downsampled (or \emph{summarized}) in increasingly larger groups using learned downsampling weights,
 and attention is calculated
 in lower resolution. The resulting 
 attention matrix has a hierarchical structure.
 FMA preserves the global receptive field of full attention
 and has an overall complexity of $\mathcal{O}(n\log n)$ or $\mathcal{O}(n)$, depending on whether the queries are
downsampled or not. It
 can be used in both autoregressive and bidirectional settings, as used, for example, in the GPT- and BERT-like language models, respectively.
We further extend FMA to two-dimensional inputs such as images.
Given a feature map of spatial size $H\times W$, 2D FMA constructs a quad-tree hierarchy:
neighboring patches attend at full resolution, whereas distant regions interact through
progressively coarser learned summaries.
Because the grouping operations factor along the two spatial axes, the computational and memory cost scales as
$\mathcal{O}(HW\log (HW))$. It can be further reduced to $\mathcal{O}(HW)$ with query downsampling, achieving linear complexity in the number of pixels.
This global yet efficient scheme eliminates the need for fixed windows or local attention heuristics employed by ViT~\cite{vit}, PVT~\cite{pvt}, or Swin~\cite{swin}, enabling megapixel scale attention without tiling. The general FMA approach is extensible to tensors with higher dimensions than 2.

Empirically, we evaluate 1D FMA on standard autoregressive and masked language-modeling benchmarks
and 2D FMA on representative image classification and semantic segmentation datasets.
FMA is compared against full attention and recent efficient alternatives, and is shown to outperform other efficient attention variants in terms of efficiency and efficacy.

 FMA inherits its divide-and-conquer spirit from the Fast Multipole Method (FMM), where near-field interactions are computed exactly and far-field effects are represented by progressively coarser multipole expansions.  
Analogously, 1D FMA computes full resolution attention for nearby tokens while summarizing distant context in a learned, hierarchical fashion. The 2D variant applies the same principle over a quad-tree of image patches.
Beyond its computational appeal, this multilevel view mirrors how humans process information: we recall recent sentences in detail but condense earlier chapters into high-level themes while reading, and when composing text we plan with coarse outlines before refining local phrasing.  
FMA formalizes this intuition, supplying language and vision models with an efficient, physics-inspired mechanism that scales to long documents and high resolution images via hierarchical context summarization.

The rest of the paper is organized as follows.
Section~\ref{sec:relatedwork} reviews the landscape of efficient attention mechanisms and situates our work within it.  
Section~\ref{sec:method} presents the Fast Multipole Attention algorithm, including both 1D and 2D formulations and implementation details.  
Section~\ref{sec:exp} describes our experimental setup and reports results on language modeling, image classification, and semantic segmentation benchmarks.  
Section~\ref{sec:discussion} discusses practical considerations, limitations, and directions for future research.

\section{Related Work}\label{sec:relatedwork}

Scaling self-attention to large input sizes has motivated a rich body of research.  
Below we group related methods and highlight the most relevant 1D and 2D variants.

\paragraph{Sparse patterns}
Local (sliding window) attention keeps only a band around the main diagonal of the
$n\times n$ attention matrix, yielding linear complexity and strong empirical results in both language~\cite{peterliu,sparse,longformer,bigbird} and vision~\cite{axial,halonet}.  
Strided or dilated patterns enlarge the receptive field by sampling farther
diagonals~\cite{sparse,longformer}.  
A complementary idea is to introduce global tokens whose representations are visible to the entire sequence
either by designating special positions within the sequence~\cite{sparse,longformer,bigbird}
or via separate summary tokens~\cite{etc,bigbird}.  
Windowed attention in images follows the same principle:
Swin Transformer~\cite{swin} and HaloNet~\cite{halonet} restrict computation to fixed-size windows
while allowing cross-window interaction through shifting or halo exchanges.

\paragraph{Clustering}
Cluster-based methods partition the tokens and compute
attention within or between clusters, leveraging the observation that
$\mathtt{softmax}$ concentrates probability mass on a few large entries.
Typical grouping rules include $k$-means~\cite{Routing,clustering-based}, locality sensitive hashing~\cite{reformer,clustered}, and iterative soft clustering as in ClusterFormer~\cite{clusterformer}.  
In 2D, axial attention~\cite{axial} may be interpreted as clustering rows or columns of an image separately, significantly reducing cost.

\paragraph{Linearization}
A large class of works replace or approximate the
$\mathtt{softmax}$ kernel with an inner product of feature maps
$\phi(q)^{\!\top}\phi(k)$, so the quadratic matrix is never formed.  
Linear Transformer~\cite{linear} uses $\phi(x)=\text{ELU}(x)+1$.
Performer~\cite{performer} and Random Feature Attention~\cite{rfa} rely on random Fourier features.
cosFormer~\cite{zhen2022cosformer} applies a cosine reweighting after $\mathtt{ReLU}$.
Although the per-token cost is $\mathcal{O}(n)$,
autoregressive training becomes sequential rather than fully parallel~\cite{tay2022efficient}.

\paragraph{Downsampling}
Memory-compressed attention~\cite{peterliu} applies stride-3 convolutions to form shorter, less detailed key and value sequences, which reduces the size of the attention matrix.
Linformer~\cite{linformer} learns a projection from
$\mathbb{R}^{n\times\hat{d}}$ to $ \mathbb{R}^{k\times\hat{d}}$, reducing the attention matrix to size
$n\times k$.  
Vision models often adopt \emph{patch merging} or pyramidal pooling to
play an analogous role, as in PVT~\cite{pvt} and SegFormer~\cite{segformer}.
Handling causal masks after downsampling, however, is nontrivial.

\bigskip
Below, we introduce several works that are particularly relevant to our method.

\paragraph{H-Transformer-1D}
H-Transformer-1D~\cite{zhu_h-transformer-1d_2021}
recursively decomposes attention into local and summarized blocks, reaching $\mathcal{O}(n)$ time and memory.
It relies on fixed averaging between levels, whereas our Fast Multipole Attention learns the basis
functions and downsampling weights end-to-end. Also, as its name suggests, H-Transformer-1D only supports 1D inputs, limiting its application. Empirically, FMA yields
higher accuracy at similar or lower cost.

\paragraph{Multi-Resolution Analysis}
Multi-Resolution Analysis (MRA) attention~\cite{zeng2022multi} adaptively refines
wherever large attention scores occur, independent of distance to the diagonal.
While theoretically appealing, the adaptive logic complicates implementation and training. MRA has so far been evaluated only on bidirectional language tasks.  
FMA instead follows the principled, physics inspired divide-and-conquer strategy of the
Fast Multipole Method~\cite{Barnes1986-ah,beatson1997short, greengard1987fast,Martinsson2015},
achieving a simpler algorithmic structure that excels on both autoregressive and bidirectional language modeling and on high resolution vision benchmarks via its 2D instantiation.

\paragraph{Swin Transformer}
Swin Transformer~\cite{swin} introduces a hierarchical design tailored for vision tasks, leveraging local self-attention within shifted, nonoverlapping windows. It partitions the input image into fixed size windows and computes attention independently within each window, thereby achieving linear computational complexity with respect to image size. Cross-window interactions are facilitated by shifting the window grid between consecutive layers, allowing gradual information exchange across the image. While highly effective, this approach attains global context only through successive stacking of multiple layers that each focuses on a different resolution, in contrast to FMA's immediate global receptive field in every layer of the Transformer network.
Our FMA 2D variant provides an isotropic $3\times3$ stencil at the
finest level and an almost isotropic $6\times6$ footprint at the first
coarse tier, independent of the query’s location.
In contrast, Swin's shifting scheme yields direction-biased,
anisotropic stencils that require many Transformer layers to propagate information
globally.  Empirically, FMA 2D attains higher accuracy than
Swin models while using less memory.

In summary, prior art either sparsifies, clusters, linearizes, or downsamples the attention matrix, each with tradeoffs in accuracy or implementation complexity.  
FMA unifies the strengths of hierarchical and physics-based fast summation, providing
learned multilevel summaries, full global context, and linear complexity in both 1D and 2D settings, and the principles of the approach are extensible to any dimension.

\section{Fast Multipole Attention}\label{sec:method}
This section presents the detailed formulation of Fast Multipole Attention. Throughout the paper we use $n$ for the sequence (or flattened image) length,
$d$ for the input embedding dimension, and $\hat{d}$ for the dimension of each
query, key, and value vector.
The integer $r\in\mathbb{N}$ denotes the base cell size for FMA. 
In all experiments we take $8\le r\le128$, but any positive integer that
divides the sequence size is admissible.
For a given level $\ell$, the index $m$ enumerates the groups produced by the
dyadic (1D) or quadtree (2D) partition.
All logarithms are base 2 and we denote $[n]$ the index set ${0,1,\ldots, n-1}$.

FMA introduces a differentiable, multi-resolution alternative to dense self-attention.  Tokens (or pixels) interact exactly with a fixed three-cell \emph{near field} (size $3r$ in 1D and $3r\times3r$ in 2D) while all farther interactions in the \emph{near field} are routed through a logarithmic hierarchy of progressively coarser groups whose representative keys and values are produced by \emph{learned} aggregation kernels.  This fast-summation scheme, inspired by the classical Fast Multipole Method, preserves the global receptive field of full attention yet reduces the self attention cost from $\mathcal{O}(n^{2})$ to $\mathcal{O}(n\log n)$ and, with query downsampling, to a strictly linear $\mathcal{O}(n)$ time and memory complexity in both sequence and image domains.

\begin{figure}[t]
  \centering
  \includegraphics[width=0.5\linewidth]{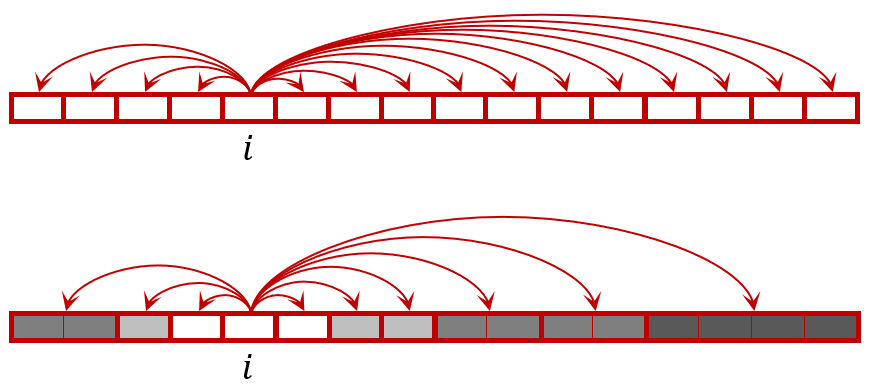}
  \caption{Conceptual view of 1D FMA.
           Top: full attention computes all $n$ pairwise interactions for
           query token $i$.  
           Bottom: FMA keeps exact attention in a $3r$ window (white cells);
           tokens beyond that window are first merged into progressively
           larger groups (in gray shades), and a single
           aggregated interaction is computed per group.}
  \label{fig:fma_concept}
\end{figure}

Figure~\ref{fig:fma_concept} illustrates the core idea behind 1D Fast Multipole Attention. It visualizes how a single query token $i$ interacts with the rest of the sequence under FMA.
The top panel shows conventional self-attention: $i$ connects to every position, producing $n$ pairwise interactions (including itself) for each token $i$ and the resulting quadratic cost.
In the bottom panel, FMA retains exact attention within a fixed
three-block neighborhood of width $3r$ (white cells), but hierarchically
aggregates increasingly distant tokens.
The first aggregation tier (light gray) merges $r$ consecutive
tokens. Outer tiers merge blocks of length
$2r$ and $4r$. Each colored block is represented by a learned summary vector, so the query
computes a single interaction per block instead of one per token.
Thus, finegrained resolution is preserved near $i$,
while long range context is captured efficiently through coarse, low rank
interactions, yielding an $\mathcal{O}(n\log n)$ global attention pattern. 

\begin{figure}[t]
  \centering
    \includegraphics[width=0.45\linewidth]{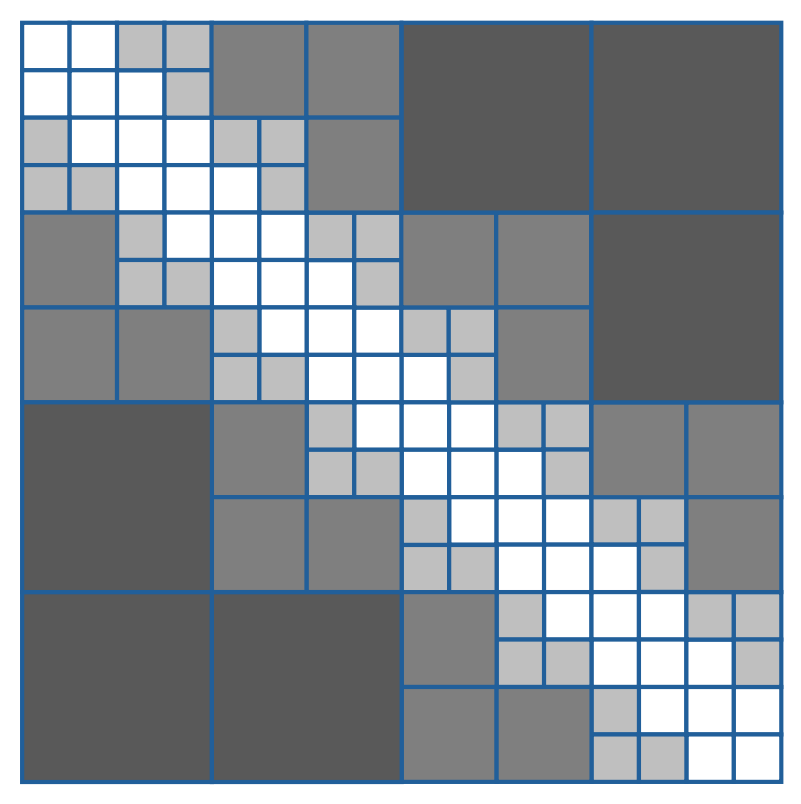}
  \caption{FMA 1D sparsified attention matrix. White cells are exact (near field). Gray cells are far field. Darker gray means coarser resolution.}
  \label{fig:fma_matrices1d}
\end{figure}

FMA replaces the dense $n\times n$ attention matrix in the Transformer architecture with a hierarchically sparsified structure that keeps exact interactions in the compact near field and represents all remaining, far-field interactions through coarse, learned summaries.  Figure~\ref{fig:fma_matrices1d} visualizes the one-dimensional FMA sparsity pattern. White cells mark the near field where every query attends to all tokens in this $3r$-wide band at full resolution.
The progressively darker gray blocks belong to successively coarser hierarchy levels that cover the far field. As the distance from the diagonal grows, tokens are aggregated into larger groups, so the attention matrix is stored and computed at ever lower resolution, reducing both memory and computation while preserving global coverage. This hierarchical block structure is a specific instance of a hierarchical matrix ($\mathcal{H}$-matrix) \cite{bebendorf2008matrices, Hackbusch1999}, a data-sparse format widely used in computational mathematics to approximate dense matrices with low-rank off-diagonal blocks.

Our exposition begins with sequences (1D FMA) and then extends the same principles to the two-dimensional case (2D FMA). In what follows we present the $\mathcal{O}(n\log n)$ version of the mechanism.

\paragraph{FMA in 1D}
Let a length $n$ token sequence be embedded in
$X\in\mathbb{R}^{n\times d}$;
linear projections produce
$Q,K,V\in\mathbb{R}^{n\times\hat d}$.
Fix a base cell size $r$ and set
$$
  L =\Bigl\lceil\log_{2}\bigl(n/r\bigr)\Bigr\rceil .
$$
Level $\ell$ $(1\le\ell<L)$ partitions the index set $[n]$
into contiguous groups
$$
  \mathcal{G}_{\ell}
  =
  \Bigl\{
     G_{\ell,m}
       =[mr2^{\ell-1},(m+1)r2^{\ell-1}) \cap [n]
  \Bigr\}_{m=0}^{M_{\ell}-1}, 
$$
where $M_{\ell}=\Bigl\lceil \tfrac{n}{r2^{\ell-1}}\Bigr\rceil$. Each group $G_{\ell,m}$ has length $r2^{\ell-1}$.
Every level learns a pair of weight vectors
$\pi^k_{\ell},\pi^v_{\ell}\in\mathbb{R}^{r2^{\ell-1}}$ that
aggregates keys and values inside a group:
\begin{equation}
  \bar k_{G_{\ell,m}}
     =\sum_{j\in G_{\ell,m}}
        \pi^k_{\ell,j-mr2^{\ell-1}}k_j,
  \qquad
  \bar v_{G_{\ell,m}}
     =\sum_{j\in G_{\ell,m}}
        \pi^v_{\ell,j-mr2^{\ell-1}}v_j .
  \label{eq:agg1d}
\end{equation}

\medskip
\noindent\emph{Near field.}
Token $i\in[n]$ lies in the base block
$b_0(i)=\lfloor i/r\rfloor$.
Its near field is the three-block window
$$
  \mathcal{N}(i)=\bigl\{j\in[n]:|b_{0}(j)-b_{0}(i)|\le1\bigr\},
  \quad
  |\mathcal{N}(i)|\le 3r .
$$

\medskip
\noindent\emph{Far field.}
For any coarse level $\ell$ let
$b_{\ell}(i)=\lfloor i/(r2^{\ell})\rfloor$.
Tokens are assigned to the far field once at the
finest level where their block is already separated from the near
field
\begin{align*}
\mathcal{F}_{\ell}(i)=
  \bigg\{
      j \in[n]: |b_{\ell}(j)-b_{\ell}(i)|\le 1,
          j\notin\mathcal{N}(i)\cup
               \bigcup_{1\leq m<\ell}\mathcal{F}_{m}(i)
  \bigg\},\\
   1\le\ell<L .
\end{align*}

The sets $\mathcal{N}(i),\mathcal{F}_{1}(i),\dots,\mathcal{F}_{L-1}(i)$
form a disjoint cover of $[n]$.

Inside $\mathcal{N}(i)$ we apply the standard softmax, and the exact attention weights inside the near field are
$$
  \alpha_{ij}=
  \frac{\exp\!\bigl(q_i^{\!\top}k_j/\sqrt{\hat d}\bigr)}
       {\displaystyle\sum_{t\in\mathcal{N}(i)}
        \exp\!\bigl(q_i^{\!\top}k_t/\sqrt{\hat d}\bigr)},
  \qquad j\in\mathcal{N}(i).
$$

For each level $\ell\geq 1$, a single interaction is computed between $q_i$ and every aggregated key $\bar k_G$ for 
$$G\in \mathcal{G}_\ell(i):= \{G_{\ell,m}: m\in [M_{\ell}],G_{\ell,m}\subseteq \mathcal{F}_{\ell}(i) \}$$
as follows:
$$
  \beta_{i,G}^{\ell}=\frac{\exp\!\bigl(q_i^{\!\top}\bar k_G/\sqrt{\hat d}\bigr)}
       {\displaystyle\sum_{G'\in\mathcal{G}_\ell(i)}
        \exp\!\bigl(q_i^{\!\top}\bar k_{G'}/\sqrt{\hat d}\bigr)}.
$$

The output token is the sum of near and far field contributions,

\begin{equation}
  \hat x_i
  =
  \sum_{j\in\mathcal{N}(i)}\alpha_{ij}v_j
  +
  \sum_{\ell=1}^{L-1}
    \sum_{G\in \mathcal{G}_\ell(i)}
      \beta_{i,G}^{\ell}\bar v_G .
  \label{eq:fma1d}
\end{equation}
\medskip
Each token attends to at most
$3r+\mathcal{O}(\log n)$ sources, yielding
$\mathcal{O}(n\log n)$ time and memory.
We note that downsampling the queries on every coarser level further reduces both to
$\mathcal{O}(n)$.

\paragraph{FMA in 2D}
Flatten a square feature map of width $W$ ($n=W^{2}$) in row major order,
yielding $X\in\mathbb{R}^{n\times d}$. Linear projections produce
$Q,K,V\in\mathbb{R}^{n\times\hat d}$.
Fix a base cell of size $3r\times 3r$ pixels and let
$$
  L =\Bigl\lceil\log_{2}\bigl(W/r\bigr)\Bigr\rceil
$$
be the depth of the quadtree hierarchy.
The feature map is partitioned into squares of side length $r2^{\ell}$,
denoted

 \begin{align*}
  \mathcal{Q}_{\ell}
  =
  \bigl\{
     S_{\ell,u,v}
       =[ur2^{\ell-1},(u{+}1)r2^{\ell-1})\times
        [vr2^{\ell-1},(v{+}1)r2^{\ell -1}) \cap [W]^2
  \bigr\}_{u,v \in [M_\ell]},\end{align*}
  where 
  $$M_\ell = \left\lceil\frac{W}{r2^{\ell -1}}\right\rceil.$$

Every level $\ell$ learns a separable kernel
$\Pi^k_{\ell}=\pi^{k,(x)}_{\ell}\otimes\pi^{k,(y)}_{\ell}, \Pi^v_{\ell}=\pi^{v,(x)}_{\ell}\otimes\pi^{v,(y)}_{\ell}
             \in\mathbb{R}^{r2^{\ell-1}\times r2^{\ell-1}}$.
For a square $S\in\mathcal{Q}_{\ell}$ with top left corner
$(x_S,y_S)$,
\begin{equation}
  \bar k_{S}=\sum_{(x,y)\in S}\!
               \Pi^k_{\ell}[x-x_S,y-y_S]k_{(x,y)},
  \quad
  \bar v_{S}=\sum_{(x,y)\in S}\!
               \Pi^v_{\ell}[x-x_S,y-y_S]v_{(x,y)}.
  \label{eq:agg2d}
\end{equation}

For pixel $i=(x_i,y_i)$ define the level $\ell$ block indices
$  B^{x}_{\ell}(i)=\lfloor x_i/(r2^{\ell})\rfloor,
  B^{y}_{\ell}(i)=\lfloor y_i/(r2^{\ell})\rfloor.$
The near field is the $3\times3$ base-cell window
$$
  \mathcal{N}(i)=
  \bigl\{j\in [W]^2:
        |B^{x}_{0}(j)-B^{x}_{0}(i)|\le1,
        |B^{y}_{0}(j)-B^{y}_{0}(i)|\le1\bigr\}.
$$

For every coarser level $\ell\ge1$ we assign each pixel $j$ to the
finest level whose square is already disjoint from the near field,
\begin{align*}
  \mathcal{F}_{\ell}(i)=
  \bigg\{
      j \in [W]^2:
      |B^{x}_{\ell}(j)-B^{x}_{\ell}(i)|\le1,
      |B^{y}_{\ell}(j)-B^{y}_{\ell}(i)|\le1,\\
      j\notin\mathcal{N}(i)\cup\bigcup_{1\leq m<\ell}\!\mathcal{F}_{m}(i)
  \bigg\}.
\end{align*}
The sets $\mathcal{N}(i), \mathcal{F}_{1}(i),\dots,\mathcal{F}_{L-1}(i)$ are pairwise
disjoint and jointly cover all pixels. Exact softmax weights within $\mathcal{N}(i)$ are
$$
  \alpha_{ij}=
    \frac{\exp\!\bigl(q_i^{\!\top}k_j/\sqrt{\hat d}\bigr)}
         {\displaystyle
          \sum_{t\in\mathcal{N}(i)}
          \exp\!\bigl(q_i^{\!\top}k_t/\sqrt{\hat d}\bigr)},
  \qquad j\in\mathcal{N}(i).
$$
For every level $\ell\ge1$ pixel $i$ attends to aggregated pixels from cells in
$\mathcal{F}_{\ell}(i)$:
$$S\in \mathcal{Q}_\ell(i):= \{S_{\ell,u,v}: u,v\in [M_{\ell}],S_{\ell,u,v}\subseteq \mathcal{F}_{\ell}(i) \}$$
$$
  \beta_{i,S}^{\ell}=
    \frac{\exp\!\bigl(q_i^{\!\top}\bar k_{S}/\sqrt{\hat d}\bigr)}
         {\displaystyle
          \sum_{S'\in\mathcal{Q}_\ell(i)}
          \exp\!\bigl(q_i^{\!\top}\bar k_{S'}/\sqrt{\hat d}\bigr)}.
$$
The resulting output feature is
\begin{equation}
      \hat x_{i}=
    \sum_{j\in\mathcal{N}(i)}\alpha_{ij}v_{j}
    +\sum_{\ell=1}^{L-1}
      \sum_{S\in\mathcal{Q}_\ell(i)}
        \beta_{i,S}^{\ell}\bar v_{S}.\label{eq:fma2d}
\end{equation}

\begin{figure}[t]
  \centering
  \begin{subfigure}{.45\linewidth}\centering
    \includegraphics[width=\linewidth]{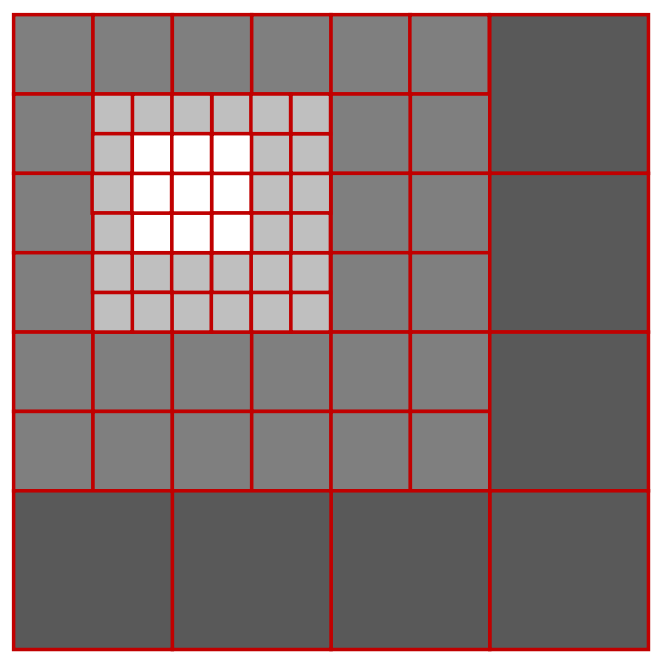}
    \caption{2D quadtree (4 levels)}
  \end{subfigure}\hfill
  \begin{subfigure}{.45\linewidth}\centering
    \includegraphics[width=\linewidth]{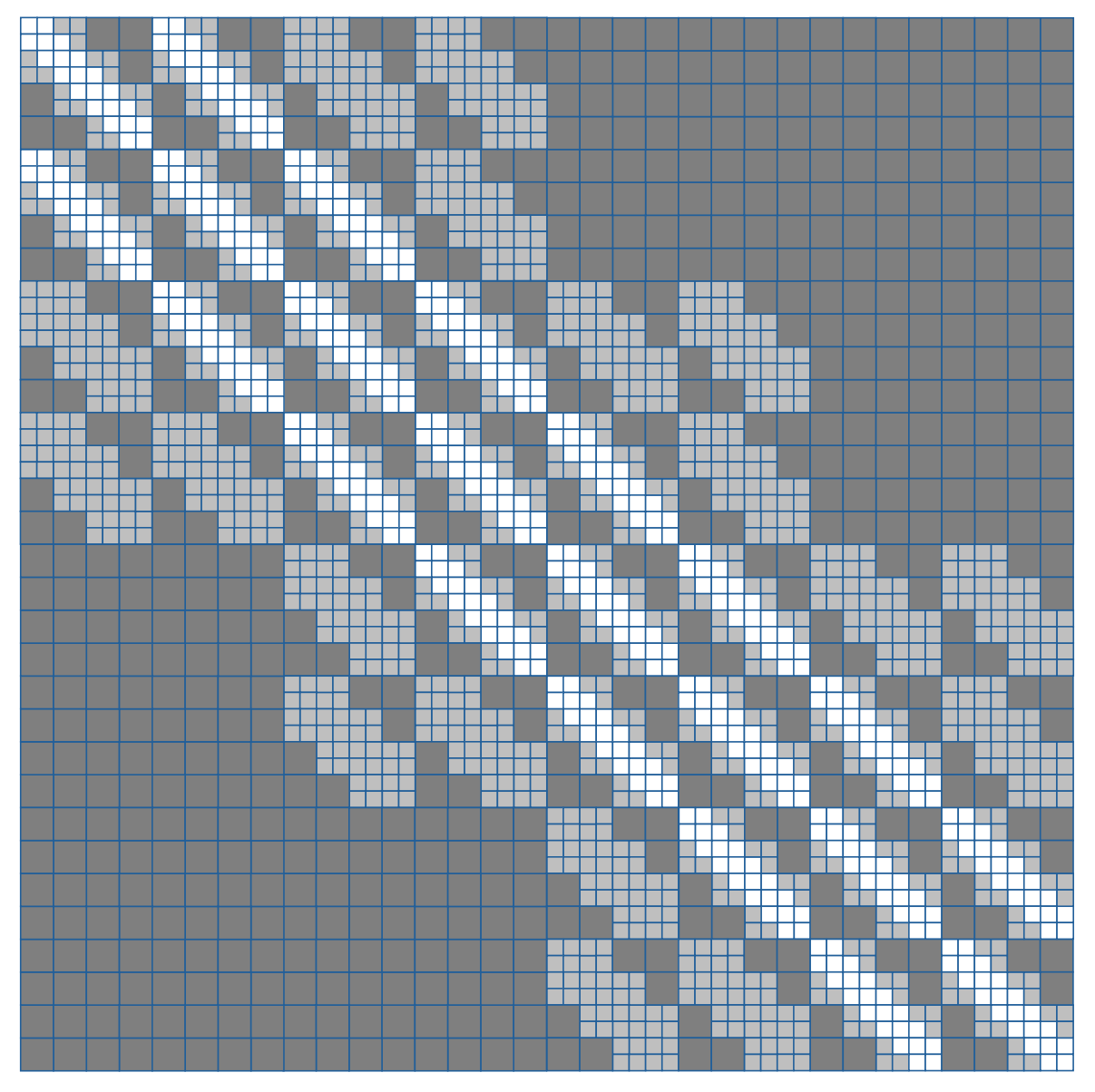}
    \caption{2D attention pattern (3 levels)}
  \end{subfigure}
  \caption{FMA 2D hierarchy and a corresponding sparsified attention matrix. White cells are exact (near field); gray cells are far field. Darker gray means coarser resolution. Panel (b) corresponds to the top left quarter of panel (a).}
  \label{fig:fma_matrices2d}
\end{figure}

Each pixel attends to at most
$9r^{2}+\mathcal{O}(\log n)$ sources, resulting in
$\mathcal{O}(n\log n)$ time and memory.
If queries are downsampled at every coarser level, both costs become linear.
The kernels $\{\pi^{(x)}_{\ell},\pi^{(y)}_{\ell}\}$ are fully differentiable
and learned end-to-end.
We visualize the quadtree partition and the
corresponding sparsified attention matrix in Figure \ref{fig:fma_matrices2d}.

\begin{figure}[t]
  \centering
  \begin{subfigure}[t]{0.4\linewidth}
    \includegraphics[width=\linewidth]{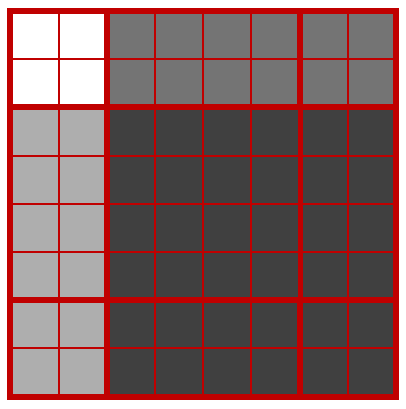}
    \caption{Stage $k$: unshifted $M\times M$ windows. (For example, the dark $4\times4$ window in the middle.)}
    \label{fig:swin_shift_a}
  \end{subfigure}\hfill
  \begin{subfigure}[t]{0.4\linewidth}
    \includegraphics[width=\linewidth]{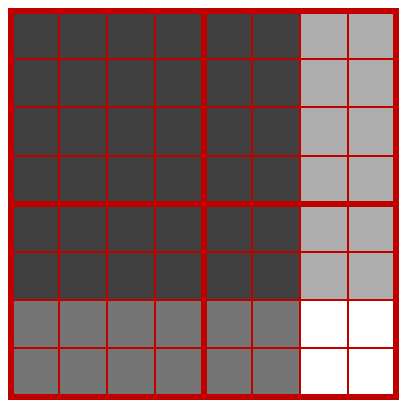}
    \caption{Stage $k{+}1$: patches shifted in the top left direction by $M/2$ pixels in a cyclic manner.}
    \label{fig:swin_shift_b}
  \end{subfigure}
  \caption{Cyclic shift mechanism in Swin Transformer. 
           Each layer computes attention in fixed local windows. The next layer
           shifts the patches in the top left direction in a cyclic manner so that boundary tokens meet new neighbors.}
  \label{fig:swin_cyclic}
\end{figure}

\paragraph{Swin Transformer}
We would like to compare our 2D FMA with the Swin Transformer, which also uses a hierarchcal design.  
Swin Transformer uses a sequence of stages with consecutively smaller numbers of pixels. Each stage uses windows with $M\times M$ patches, and has multiple Transformer layers.
The first layer in a stage computes attention exactly for pixels within the same nonoverlapping $M\times M$ window. Subsequent layers compute attention for shifted windows. For example, the second layer may compute attention in $M\times M$ windows composed of patches shifted in the top left diagonal direction by $\lfloor M/2\rfloor$ pixels in a cyclic manner,
bringing previously separated tokens into the same window, see Figure~\ref{fig:swin_cyclic}. The shifting is implemented in a cyclic manner for efficiency reasons.
The shifts provide limited cross-window communication while producing
direction-biased, anisotropic stencils.

\begin{figure}[t]
  \centering
  \begin{subfigure}[t]{0.45\linewidth}
    \includegraphics[width=\linewidth]{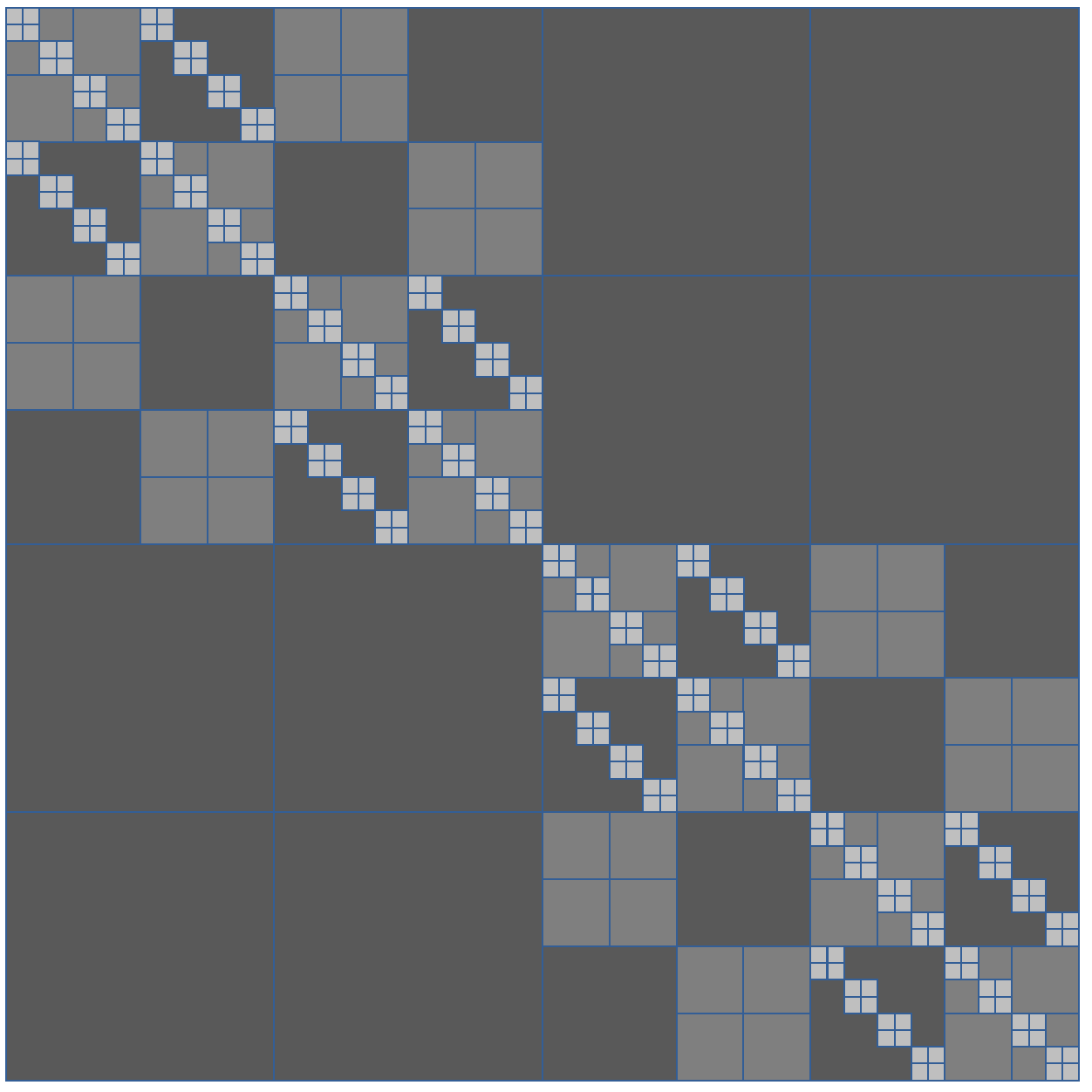}
    \caption{Swin block-diagonal pattern: local $M\times M$ windows in the first unshifted layer of each stage.}
    \label{fig:swin_local}
  \end{subfigure}\hfill
  \begin{subfigure}[t]{0.45\linewidth}
    \includegraphics[width=\linewidth]{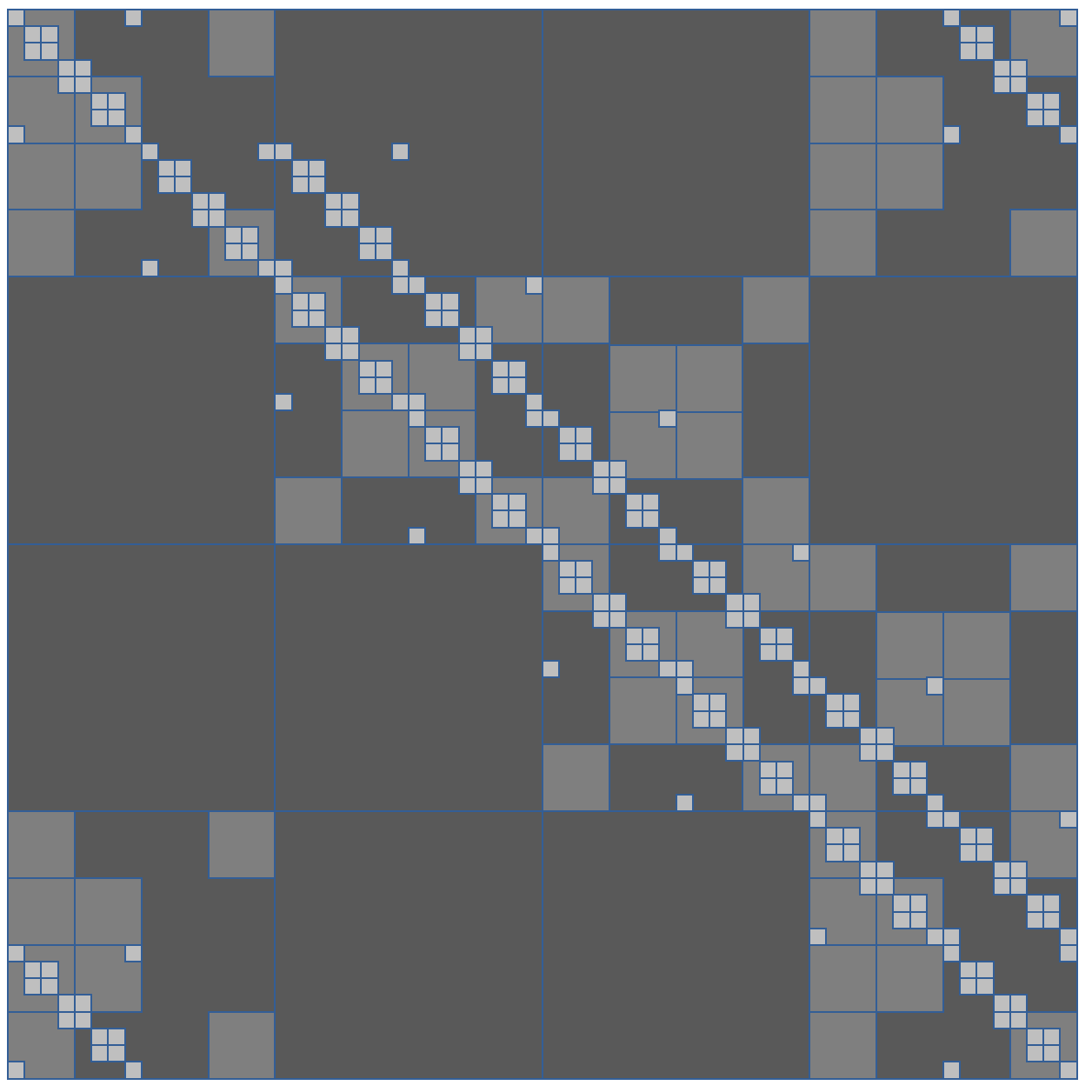}
    \caption{Shifted windows at the next pass allow cross-window exchange along stitched diagonals.}
    \label{fig:swin_shift}
  \end{subfigure}
  \caption{Attention sparsity across all three stages of a Swin Transformer.
           cells with different shades belong to different stages. Cells may overlap. Only the lightest (i.e., finest) shade is shown when cells overlap.  
           Each layer attends only within a fixed window, so global information must percolate through successive layers with alternating shifts, unlike the single layer global reach of FMA2D (Fig.~\ref{fig:fma_matrices2d}).}
  \label{fig:swin_attention}
\end{figure}

Figure~\ref{fig:swin_attention} overlays the sparsity patterns from several
stages of a Swin Transformer with different resolutions, coloring each stage differently.  
Figure~\ref{fig:swin_local} shows the local connections from the first layer of each stage, without shifting.
Because every layer is local, the union of the first layers over all stages gradually forms diagonal
bands. More complete global coverage appears only after adding shifted layers to each stage, see Figure~\ref{fig:swin_shift}.  
FMA, in contrast, does not use multiple stages and provides a global receptive field in every individual Transformer layer: every layer uses a multilevel representation with a fixed $3r\times3r$ exact near field plus learned low-rank summaries for the far field. The FMA2D approach also results in more isotropic global connectivity; compare Figure~\ref{fig:swin_local} and Figure~\ref{fig:swin_shift} with Figure~\ref{fig:fma_matrices2d}.  
This divide-and-conquer design yields higher accuracy at both
$224^{2}$ and $384^{2}$ resolutions (Table~\ref{tab:imagenet}) while
reducing peak GPU memory by 0.7-2.9 GB, confirming the efficiency of 2D FMA’s
global yet hierarchical attention in each layer.

\paragraph{Coarse level rank $p$}  
FMA can represent each 1D or quadtree (2D) group with a
rank-$p$ summary rather than a single vector.
Specifically, every level $\ell$ learns a set of ``basis'' functions
$\{\pi^{(s)}_{\ell}\in\R^{2^{\ell}}\}_{s=1}^{p}$ in 1D
(or a separable pair
$\{\Pi^{(s)}_{\ell}
       =\pi^{(x,s)}_{\ell}\otimes\pi^{(y,s)}_{\ell}\}_{s=1}^{p}$ in 2D).
A group $G$ then stores \emph{p} aggregated keys and values,
$$
  \bar k^{(s)}_{G}=\sum_{j\in G}\pi^{k,(s)}_{\ell,j}k_j,
  \qquad
  \bar v^{(s)}_{G}=\sum_{j\in G}\pi^{v,(s)}_{\ell,j}v_j,
  \qquad s=1,\dots,p,
$$
yielding a $p\times\hat{d}$ low-rank factorization for the far-field block.
During attention, the query $q_i$ contracts with the
$p$ representatives $\{\bar k^{(s)}_{G}\}$. The resulting scores are
normalized across all groups on that level exactly as in
Eqs.~\eqref{eq:fma1d}-\eqref{eq:fma2d}.
Setting $p{=}1$ recovers the default configuration used in
Section \ref{sec:method}. Larger $p$ increases modeling capacity at the
cost of a proportional runtime and memory factor, as explored in
Section \ref{ablation}.

\paragraph{Implementation} Our implementation realizes the theoretical speedup of FMA with a custom TVM \cite{tvm} schedule and optimized CUDA kernels. The kernels implement three computational stages: exact near-field attention, hierarchical interactions of keys and queries, and weighted combination of all values. Crucially, the low-rank far-field blocks that appear conceptually in the methodological exposition are never materialized in device memory. Instead, each block’s contribution is generated on the fly from its learned group representatives. By avoiding explicit construction of any off-diagonal submatrices, our implementation maintains the theoretical complexity of $\mathcal{O}(n\log n)$ or $\mathcal{O}(n)$ in both runtime and memory, while fully exploiting GPU parallelism.

\begin{figure}[h!]
\centering
\includegraphics[width=0.9\linewidth]{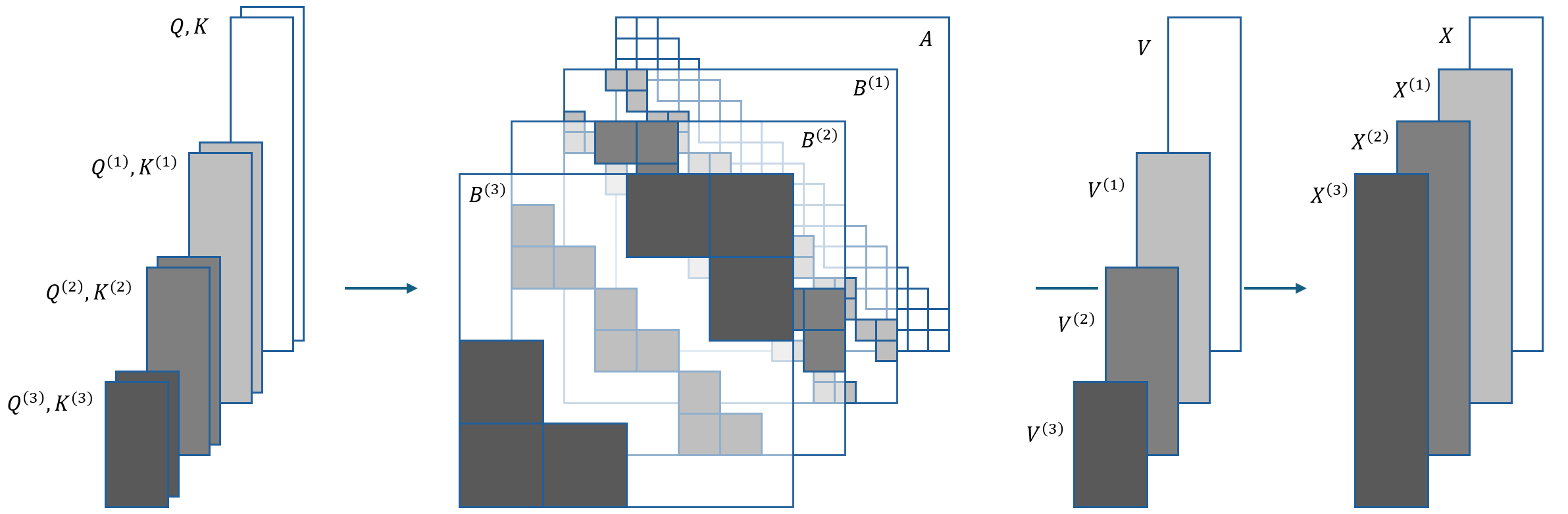}
\caption{Overview of the 1D FMA implementation. Queries and keys are projected and aggregated hierarchically without explicitly materializing low rank off-diagonal blocks. Darker blocks indicate coarser resolution levels.}
\label{fig:fma_flowchart}
\end{figure}

\begin{figure}[h!]
  \centering
  \begin{subfigure}[b]{0.9\columnwidth}
    \centering
    \includegraphics[width=0.8\linewidth]{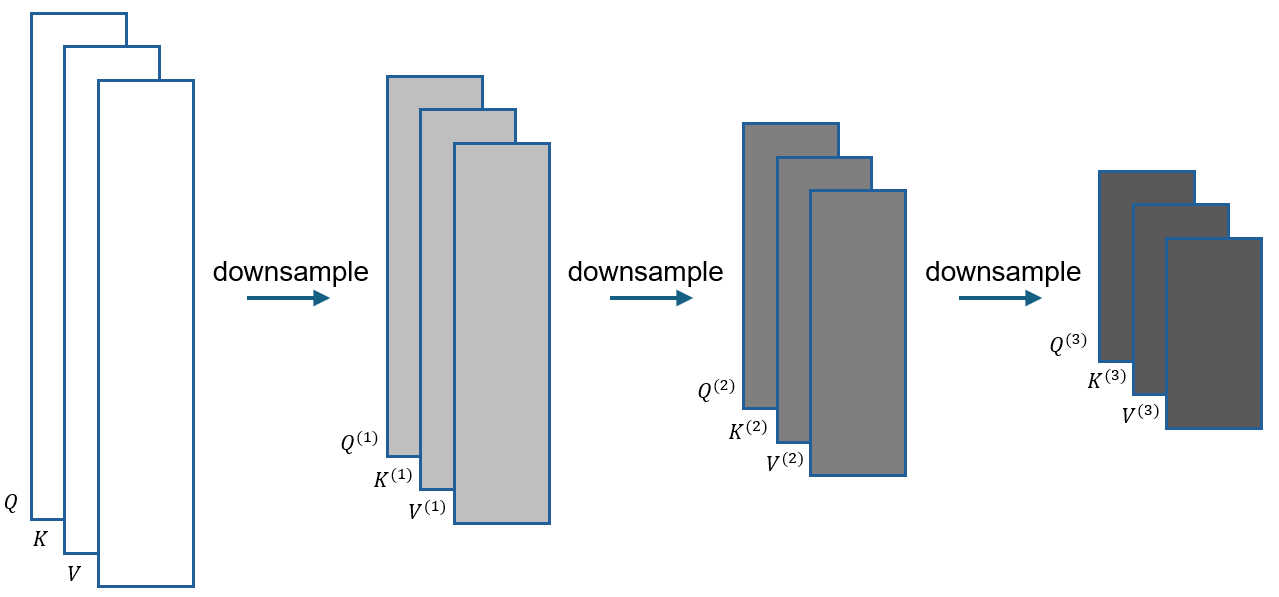}
    \caption{Down-sampling across levels.}
    \label{fig:overview1}
  \end{subfigure}\par\medskip
  \begin{subfigure}[b]{0.9\columnwidth}
    \centering
    \includegraphics[width=\linewidth]{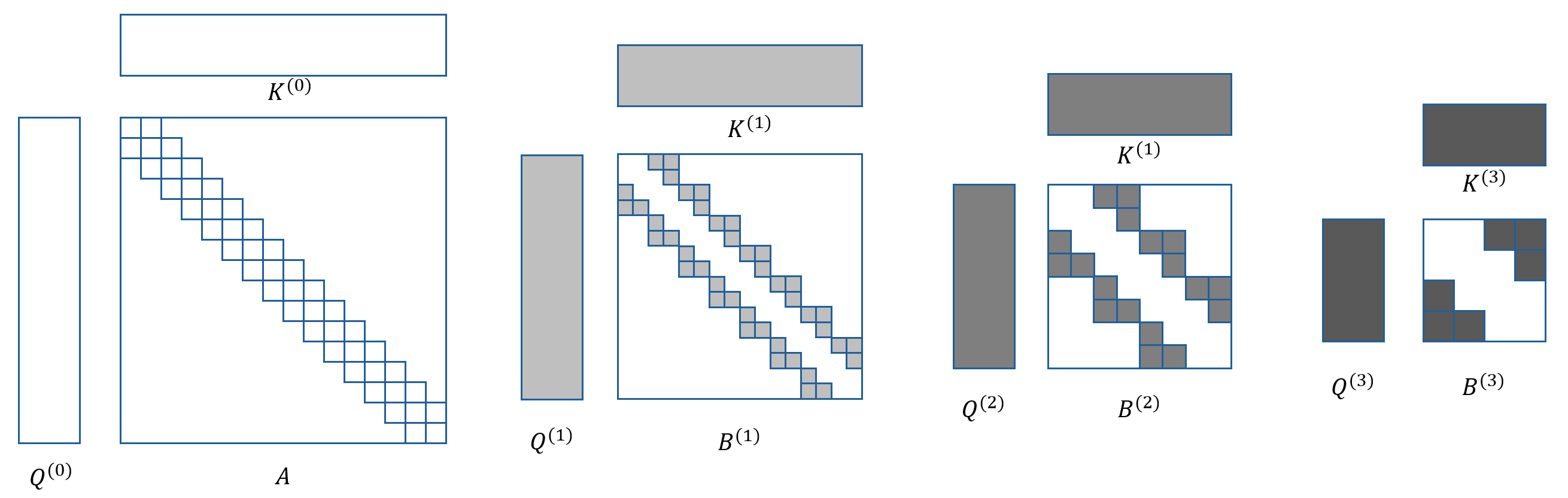}
    \caption{Level-wise blocks for $Q,K$ and banded sub-matrices.}
    \label{fig:overview2}
  \end{subfigure}\par\medskip
  \begin{subfigure}[b]{0.9\columnwidth}
    \centering
    \includegraphics[width=\linewidth]{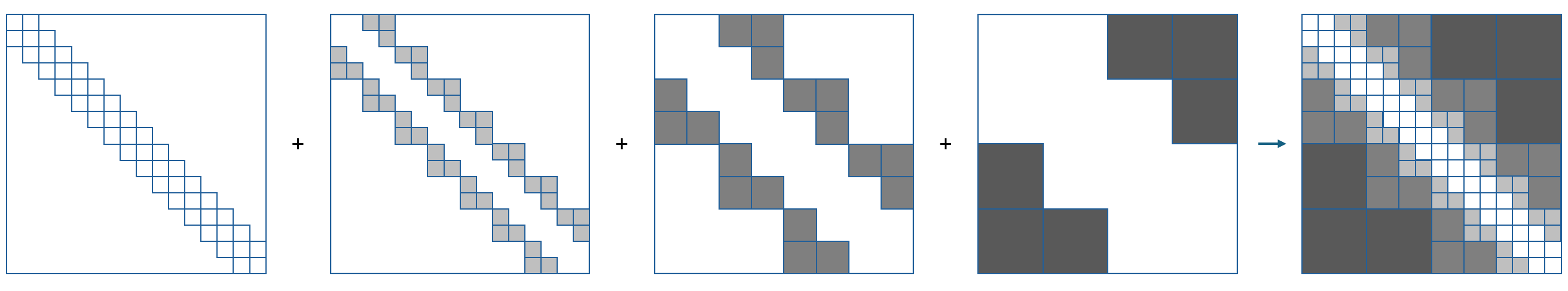}
    \caption{Superposition of near/far blocks forms the final sparse matrix.}
    \label{fig:overview3}
  \end{subfigure}
  \caption{Overview of bidirectional FMA with four hierarchy levels.
           (a) Multilevel down-sampling of $Q,K,V$;
           (b) hierarchically banded sub-matrices $A,B^{(\ell)}$ containing
           weights $\alpha,\beta$ (Sec.~\ref{sec:method});
           (c) their superposition yields the final hierarchical attention
           matrix.}
  \label{fig:fma_flowchart1}
\end{figure}

Figure~\ref{fig:fma_flowchart} provides an overview of this efficient implementation for the 1D Fast Multipole Attention. At each layer $\ell$, queries $ Q^{(\ell)} $ and keys $ K^{(\ell)} $ are first generated through learned linear projections from input representations. Instead of forming the dense attention matrix explicitly, attention computations follow the hierarchical sparsity pattern imposed by FMA. Exact attention weights are computed within small diagonal blocks, while distant interactions are progressively aggregated into low rank summaries using learned aggregation kernels. Thus, the intermediate attention matrix exhibits a hierarchical block structure, with exact computations at fine resolution, stored as a matrix $A$ in sparse form and low rank off-diagonal blocks, stored as sparse matrices $B^{(\ell)}$. Figure~\ref{fig:fma_flowchart1} demonstrates how the hierarchical attention matrix is obtained.

Finally, the values $ V^{(\ell)} $ are combined efficiently using the sparsified attention scores to yield output features $ X^{(\ell)} $. This implementation strategy leverages GPU hardware effectively with specialized CUDA kernels to deliver the theoretical benefits of Fast Multipole Attention.

\section{Experiments}\label{sec:exp}

This section benchmarks Fast Multipole Attention in both one-dimensional
and two-dimensional settings and analyzes its memory behavior.
Two language modeling tasks serve as 1D tests, while large scale image
classification and segmentation cover the 2D regime. FMA means FMA with $\mathcal{O}(n\log n)$ complexity and FMA-linear has $\mathcal{O}(n)$ complexity with query downsampling. The tasks we consider are:

\begin{itemize}
  \item Autoregressive language modeling on enwik8
        (Sec.~\ref{exp1}).
  \item Masked language modeling on WikiText-103
        (Sec.~\ref{exp2}).
  \item ImageNet-1K classification at $224^{2}$ and $384^{2}$
        resolutions (Sec.~\ref{exp3}).
  \item ADE20K semantic segmentation (Sec.~\ref{exp4}).
\end{itemize}

Finally, we compare the memory and time efficiency of FMA to other efficient attention variants. For the 1D experiments we compare against six strong efficient-attention
methods: Memory-Compressed Attention (MCA)~\cite{peterliu},
Reformer~\cite{reformer}, Linear Transformer~\cite{linear},
BigBird~\cite{bigbird}, H-Transformer-1D~\cite{zhu_h-transformer-1d_2021},
and Multi-Resolution Analysis (MRA) attention~\cite{zeng2022multi}.
Context lengths vary from 512 to 4096 tokens, and every variant uses the
same 6-layer decoder (embedding 768, 12 heads, pre-norm, learned
positional embeddings). For the 2D tasks we benchmark against
ViT-B/16~\cite{vit}, Swin-B/L~\cite{swin}, and Segformer~\cite{segformer}.

In language training, hyperparameters are tuned on full attention and then frozen for each
efficient substitute.
In the autoregressive task, the learning rate linearly warms to $6\times10^{-4}$ over 200 steps, then
decays cosinely to $6\times10^{-5}$ over 60k updates.
We use Adam ($\beta_1{=}0.9$, $\beta_2{=}0.98$), weight decay~0.4,
dropout~0.3, and global batch size~64.
All experiments are run on NVIDIA A100 or H100 GPUs via \texttt{fairseq}~\cite{fairseq}.
FMA has two hyperparameters: base cell size $r$ and coarse rank $p$.
Unless noted, $p=4$,
$r=64$ for $n\le 1024$ and $r=128$ otherwise, yielding
$\lceil\log_2(n/r)\rceil$ levels.
Larger $r$ or $p$ improves accuracy at extra cost
(Sec.~\ref{ablation}).  
Reformer uses two LSH rounds with $\sqrt n$ buckets rounded up to a
power of two. H-Transformer fixes its numerical rank to 16.

Image transformer models follow the Swin approach: 300 epochs, AdamW, and cosine LR peaking at $1\times10^{-3}$. Data augmentation methods like Mixup and CutMix are used.
ADE20K models train for 160k iterations on $512^{2}$ crops following SegFormer.
Peak GPU memory is measured on one Nvidia A100 GPU. Quadratic baselines that exceed
device limits use gradient accumulation, and the accumulated footprint is
reported in the tables (grey rows denote $\mathcal{O}(n^{2})$ attention).

We report four standard metrics.
Bits per character (bpc) quantifies the negative log-likelihood of a
character-level language model in bits, where lower value is better.
For word level models we use perplexity (PPL), the exponentiated average
negative log-likelihood, where lower values again indicate superior modeling
capacity.
Image classification performance is measured by top-1 accuracy, the
percentage of validation images whose predicted class matches the ground
truth.
Semantic segmentation quality is reported as mean Intersection over Union (mIoU), the average Jaccard index across all classes. Higher mIoU reflects more precise pixel predictions.

\subsection{Autoregressive Language Modeling}\label{exp1}

\begin{table}[t]
\footnotesize\centering
\begin{tabular}{lccc}
\toprule
Model & Context $n$ & Test bpc $\downarrow$ & Memory (GB)\\
\midrule
\rowcolor[gray]{.92} Full (quadratic) & 512  & 1.346 & 16.5 \\
\rowcolor[gray]{.92} MCA (quadratic)  & 512  & 1.439 & 13.5 \\
Reformer           & 512  & 1.372 & 15.5 \\
Linear             & 512  & 1.424 & 38.2 \\
BigBird            & 512  & 1.465 & 16.4 \\
H-Transformer      & 512  & 1.853 & 12.3 \\
FMA        & 512  & \textbf{1.353} & \textit{11.6} \\
FMA-linear     & 512  & \textit{1.391} & \textbf{9.8} \\
\midrule
\rowcolor[gray]{.92} Full (quad.)      & 1024 & 1.236 & 46.5 \\
\rowcolor[gray]{.92} MCA (quad.)       & 1024 & 1.351 & 27.4 \\
Reformer            & 1024 & \textit{1.271} & 31.8 \\
Linear              & 1024 & 1.343 & 74.9 \\
BigBird             & 1024 & 1.345 & 23.1 \\
H-Transformer       & 1024 & 1.789 & 23.7 \\
FMA        & 1024 & \textbf{1.256} & \textit{21.6} \\
FMA-linear      & 1024 & 1.298 & \textbf{17.6} \\
\midrule
\rowcolor[gray]{.92} Full (quad.)      & 2048 & 1.128 & $2\times71.5$ \\
\rowcolor[gray]{.92} MCA (quad.)       & 2048 & 1.276 & 73.4 \\
Reformer   &2048&1.252&62.9\\
Linear      &2048&1.274 &2$\times$72.1\\
BigBird     &2048&1.281&39.5\\
H-Transformer       & 2048 & 1.684 & \textit{38.4} \\
FMA        & 2048 & \textbf{1.220} & 48.2 \\
FMA-linear      & 2048 & \textit{1.244} & \textbf{34.1} \\
\midrule
\rowcolor[gray]{.92} Full (quad.)      & 4096 & 1.091 & $8\times61.7$ \\
\rowcolor[gray]{.92} MCA (quad.)      &4096&1.212& $4\times 49.2$\\
Reformer   &4096&1.180& $2\times 63.6$\\
Linear      &4096&1.207&4$\times$71.0\\
BigBird     &4096&1.216&72.2\\
H-Transformer       & 4096 & 1.535 & \textbf{66.0} \\
FMA         & 4096 & \textbf{1.133} & 2$\times$ 50.5 \\
FMA-linear     & 4096 & \textit{1.154} & \textit{69.4} \\
\bottomrule
\end{tabular}
\caption{Character-level language modelling on enwik8. 
Lower bits-per-character (bpc) is better.  Grey rows denote quadratic baselines.  Bold figures mark the best efficient ($<\!\mathcal{O}(n^{2})$) method per block. Second-best is indicated by italics. Memory is the peak GPU consumption. Some models require gradient accumulation, hence multiple passes (e.g.\ “$2\times$”).}
\label{tab:enwik8}
\end{table}

Table~\ref{tab:enwik8} shows that the FMA variant with $\mathcal{O}(n\log n)$ complexity outperforms the other efficient transformers in terms of accuracy. When the sequence length increases, the gap over the next-best efficient model (Reformer) increases (1.4\%, 1.2\%, 2.6\%, and 4.1\% for $n=512,1024,2048$ and $4096$, respectively). 
Our $\mathcal{O}(n\log n)$ variant stays in the top 4 lowest memory among all efficient transformers while our linear variant is best or second-best.

Full attention remains most accurate but becomes infeasible beyond 2048 tokens. Compared to the $\mathcal{O}(n\log n)$ variant, FMA-linear trades a modest loss in bpc for additional memory savings.   FMA and FMA-linear both scale to 4096 with the best accuracy, and excellent memory performance for FMA-linear.  Compared with H-Transformer, FMA-linear achieves dramatically lower bpc and comparable memory. Overall, these results demonstrate that both FMA variants offer a superior combination of accuracy and scalability.

\begin{figure}[h]
    \centering
    \begin{subfigure}[b]{0.3\textwidth}
        \centering
        \includegraphics[width=\linewidth]{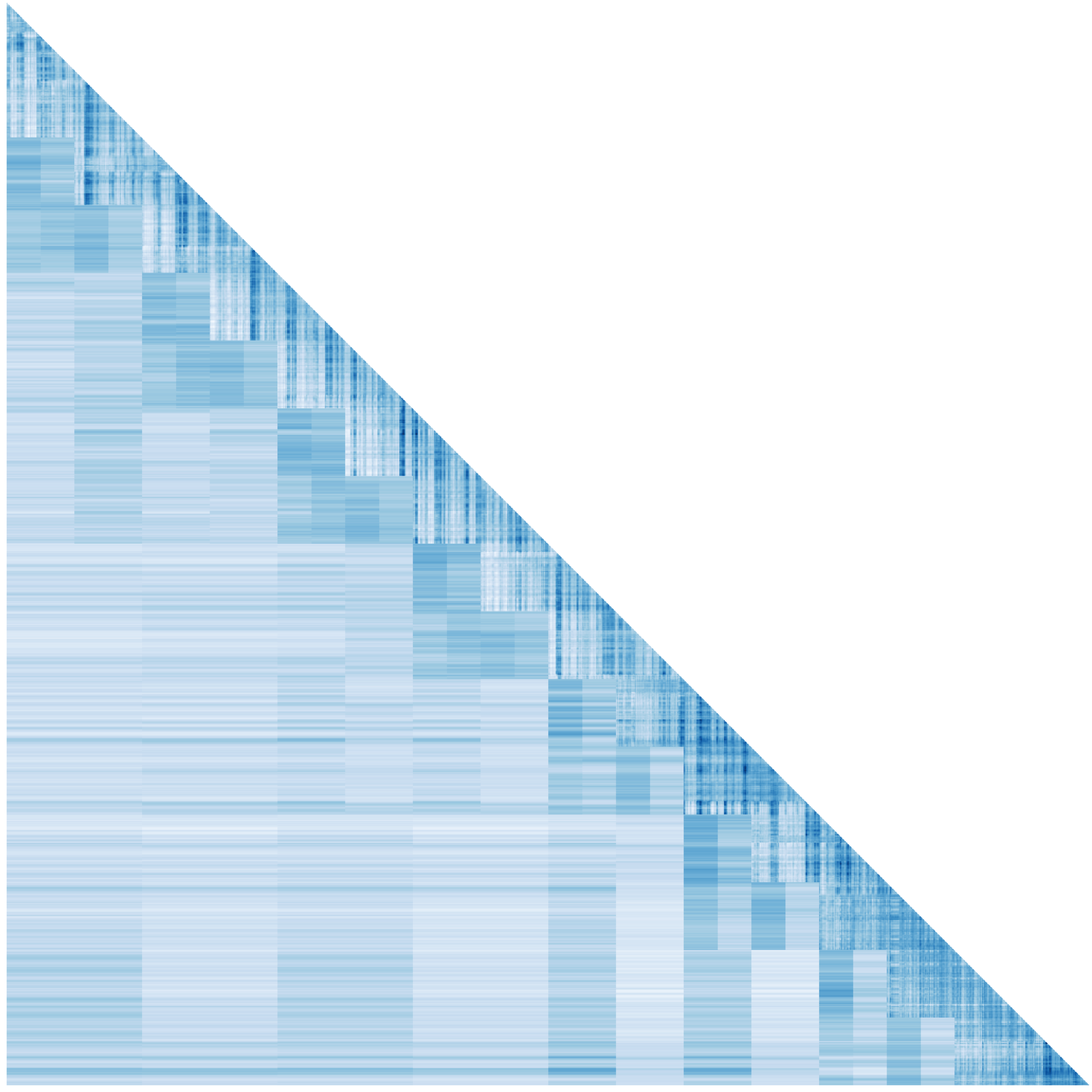}
    \end{subfigure}
    \hfill
    \begin{subfigure}[b]{0.3\textwidth}
        \centering
        \includegraphics[width=\textwidth]{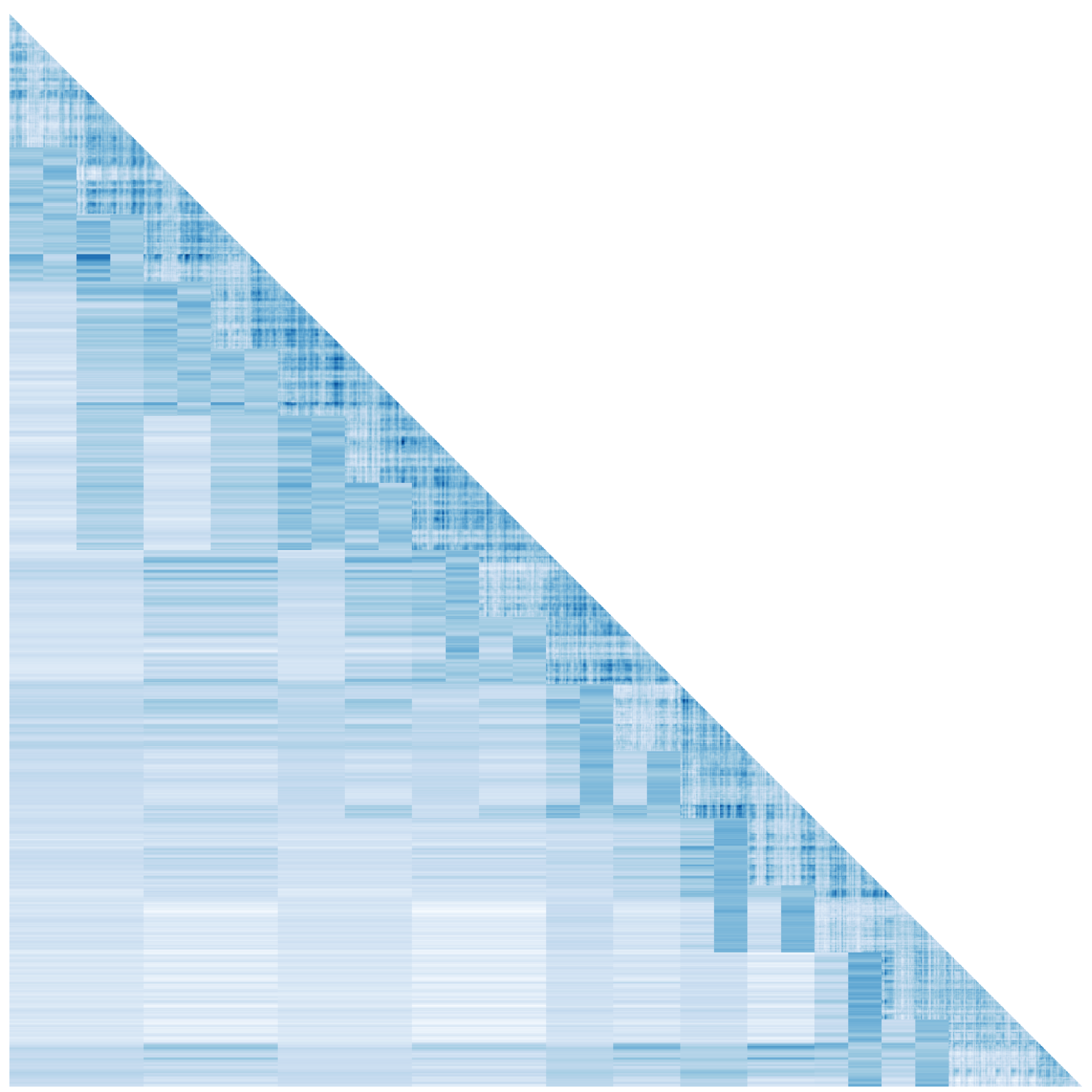}
    \end{subfigure}
    \hfill
    \begin{subfigure}[b]{0.3\textwidth}
        \centering
        \includegraphics[width=\textwidth]{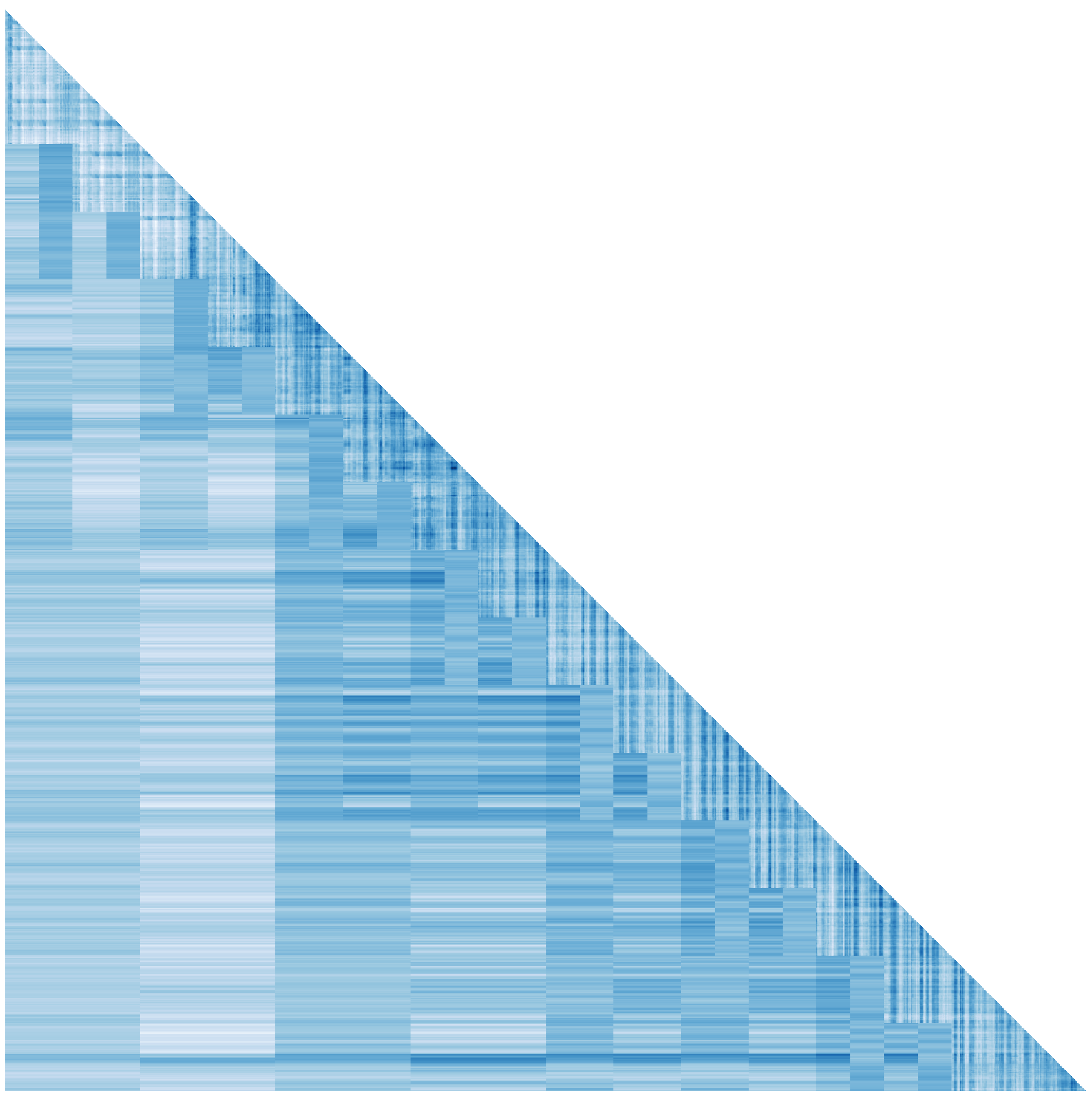}
    \end{subfigure}
    \caption{Some sample FMA attention matrices from various layers and heads for the autoregressive task of Table 1 with $n=1024$. The left, middle, and right panels represent a head from layer 2, 3 and 4, respectively.}
    \label{fig:attn}
\end{figure}
Figure~\ref{fig:attn} shows sample attention maps from a trained FMA model, confirming the divide-and-conquer strategy.  As expected, high magnitude coefficients cluster near the diagonal where FMA allocates full resolution. And some important long range dependencies are captured by the hierarchically coarsened far-field groups.

\subsection{Bidirectional Language Modeling} \label{exp2}

\begin{table}[htbp]
\footnotesize
\centering
\begin{tabular}{l c c c } 
 \toprule
 model  &\makecell{context\\size $n$}& ppl(test)$\downarrow$ & \makecell{memory\\footprint}\\ 
 \midrule
\rowcolor[gray]{0.9} Full  & 512 & 9.21 & 15.4 GB\\
  Reformer  & 512 & 11.03 & 14.6 GB\\
   Linear  & 512 & 14.95 & 11.8 GB\\
  MRA  & 512 & 11.82  & 11.1 GB\\
  H-Transformer  & 512 &  10.60 & 10.7 GB\\ 
  FMA  & 512 & \textbf{9.45} & 10.9 GB\\
  FMA-linear  & 512 & 10.78 & \textbf{8.2 GB}\\
 \cmidrule(r){1-4}
\rowcolor[gray]{0.9} Full & 1024 & 8.89 & 36.8 GB\\
  Reformer  & 1024 & 10.56  & 27.4 GB\\
   Linear  & 1024 & 14.24 & 22.0 GB\\
  MRA  & 1024 &  10.20 & 17.5 GB\\
  H-Transformer  & 1024 &  9.81 & 16.9 GB\\
  FMA & 1024 &  \textbf{9.06} & 18.6 GB\\
  FMA-linear & 1024 &  10.13 & \textbf{15.6 GB}\\ 
 \cmidrule(r){1-4}
\rowcolor[gray]{0.9} Full & 2048 & 8.70 & 2$\times$56.5 GB\\
  Reformer & 2048 &  9.90 & 49.2 GB\\
  Linear & 2048 & 14.06 & 48.1 GB\\
    MRA & 2048 &  9.66 & 35.3 GB\\
  H-Transformer & 2048 &  9.45 & 33.1 GB\\
  FMA & 2048 &  \textbf{8.95} & 37.0 GB\\
  FMA-linear & 2048 &  9.64 & \textbf{29.5 GB}\\
    \bottomrule\\
\end{tabular}
\caption {\label {ppl}Bidirectional (masked) language modeling results on Wikitext-103. Models with quadratic complexity are highlighted in gray. Bold numbers are best among the efficient transformers (not counting the $\mathcal{O}(n^2)$ transformers). With the same context size, our FMA obtains the best accuracy and uses less memory compared to other efficient attention variants.}
\end{table}
We evaluate our model on the Wikitext-103 dataset, a common benchmark for testing long-term dependencies in language models. We train a bidirectional Transformer encoder (6 layers, 12 heads, embedding dimension of 768) using a masked language modeling objective, randomly masking 15\% of the input tokens. To isolate the impact of the attention mechanism, we replace the standard full self-attention with FMA and other efficient variants, keeping the rest of the architecture constant.

The results in Table \ref{ppl} demonstrate that the FMA family establishes a superior balance between accuracy and efficiency. The standard FMA variant consistently achieves the best accuracy among all efficient methods, with its perplexity of 8.95 at a context size of 2048 nearly matching the quadratic full attention baseline (8.70). Meanwhile, the FMA-linear variant offers a compelling trade-off, reducing the memory footprint to 29.5 GB while maintaining highly competitive performance. Together, these results demonstrate that FMA provides performance comparable to full attention without incurring its prohibitive computational cost.

\subsection{Image Classification}\label{exp3}

To demonstrate that the 2D formulation of Fast Multipole Attention (FMA2D) performs effectively for images, we evaluate it on the ImageNet-1K benchmark (1.28M training images, 1K classes). The FMA2D-B and FMA2D-L models are obtained by replacing each standard attention block in a ViT backbone  (ViT-B/16 or ViT-L/16) with our 2D FMA operator introduced in Section~\ref{sec:method}. All other hyperparameters, including hidden size (768 for ViT-B, 1024 for ViT-L), MLP dimensionality (3072 for ViT-B, 4096 for ViT-L), and layer count (12 for ViT-B, 24 for ViT-L), remain unchanged. We train for 300 epochs with a cosine learning-rate schedule peaking at $1\times10^{-3}$, following the procedure outlined in~\cite{vit}. We set $r=4$ for image size 224 and $r=8$ for image size $384$.

\begin{table}[h]
\footnotesize\centering
\begin{tabular}{lcccc}
\toprule
\multirow{2}{*}{Model} & \multirow{2}{*}{Params} & Img & Top-1 & Memory \\
                       &                         & size & (\%)↑ & (GB)↓ \\
\midrule
ViT-B/16~\cite{vit}      & 86M & 224 & 81.8 & 17.2 \\
Swin-B~\cite{swin}       & 88M & 224 & 83.5 & 15.3 \\
FMA2D-B      & 92M & 224 & \textbf{84.0} & \textbf{14.6} \\
\midrule
ViT-B/16                 & 86M & 384 & 83.6 & 41.4 \\
Swin-B                   & 88M & 384 & 84.5 & 28.1 \\
FMA2D-B          & 85M & 384 & \textbf{85.3} & \textbf{26.6} \\
\midrule
ViT-L/16                 & 307M & 384 & 85.2 & 103.5 \\
Swin-L                   & 197M & 384 & 87.0 & 66.8 \\
FMA2D-L        & 201M & 384 & \textbf{87.5} & \textbf{63.4} \\
\bottomrule
\end{tabular}
\caption{Single-crop top-1 accuracy and peak training memory on ImageNet-1K. FMA2D matches the parameter counts of the ViT and Swin baselines yet consistently achieves higher accuracy with lower GPU memory consumption at the same resolutions.}
\label{tab:imagenet}
\end{table}

\begin{table}[h]
\footnotesize\centering
\begin{tabular}{lcccc}
\toprule
\multirow{2}{*}{Model} & \multirow{2}{*}{Params} & Img & Top-1 & Memory \\
                       &                         & size & (\%)↑ & (GB)↓ \\
\midrule
ViT-B/16     & 86M & 384 & 84.0 & 41.4 \\
Swin-B     & 88M & 384 & 86.4 & 28.1 \\
FMA2D-B         & 85M & 384 & \textbf{87.1} & \textbf{26.6} \\
\midrule
ViT-L/16                 & 307M & 384 & 85.8 & 103.5 \\
Swin-L                   & 197M & 384 & 88.1 & 66.8 \\
FMA2D-L          & 201M & 384 & \textbf{88.8} & \textbf{63.4} \\
\bottomrule
\end{tabular}
\caption{Single-crop top-1 accuracy and peak training memory on ImageNet-1K with ImageNet-22K pretrain.}
\label{tab:imagenetpretrain}
\end{table}
Table~\ref{tab:imagenet} demonstrates that FMA2D models outperform corresponding VIT models by a large margin in both 224 and 384 resolution with fewer parameters. They also outperform Swin-B and Swin-L with similar parameter count.

The results in Table~\ref{tab:imagenet} demonstrate that FMA2D models achieve substantially higher accuracy than corresponding ViT models at both 224 and 384 resolutions while using fewer parameters. Furthermore, FMA2D outperforms Swin-B and Swin-L at a similar parameter scale. When pretrained on ImageNet-22K (Table~\ref{tab:imagenetpretrain}), all models show a general improvement in accuracy. The same performance trends persist, with FMA2D models improving over the existing methods.

Because FMA2D never explicitly forms off-diagonal attention blocks, its GPU footprint grows significantly slower than quadratic attention architectures. This enables efficient training on inputs such as $384\times 384$ or larger that would otherwise exceed practical GPU memory limits.

\subsection{Semantic Segmentation}\label{exp4}

To evaluate Fast Multipole Attention on dense prediction, in this comparison we pretrain every model on ImageNet-22K and finetune on the ADE20K benchmark (150 classes, 25k images). FMA2D-B-SegFormer and FMA2D-L-SegFormer are obtained by replacing the self-attention blocks in SegFormer-B4 and B5 with 2D FMA. The overall network depth, width, and MLP channels remain unchanged.

To ensure a fair comparison, all models were fine-tuned using a consistent training recipe. We trained for 160,000 iterations on $512\times512$ image crops. The models were optimized using AdamW with a weight decay of 0.01 and an initial learning rate of $6\times 10^{-5}$, which followed a polynomial decay schedule. Data augmentation techniques including RandAugment, Mixup, and CutMix, were applied. Finally, predictions were generated using a sliding window inference strategy.

\begin{table}[h]
\footnotesize\centering
\begin{tabular}{lcccc}
\toprule
Method  & Params & mIoU (\%)↑\\
\midrule
Swin-B-UPerNet & 121M & 48.1  \\
Swin-L-UPerNet & 234M & 50.5  \\
SegFormer-B4    &  64M & 51.2  \\
SegFormer-B5    &  85M & 51.5  \\
FMA2D-B-SegFormer &71M & 51.9\\
FMA2D-L-SegFormer &90M & \textbf{52.4}\\
\bottomrule
\end{tabular}
\caption{ADE20K validation results with ImageNet pretraining.  
FMA2D raises single scale mIoU over each SegFormer baseline and even surpasses a much larger UPerNet+Swin-L model.}
\label{tab:ade20k_scratch}
\end{table}

As shown in Table~\ref{tab:ade20k_scratch}, FMA2D-L-SegFormer achieves the highest accuracy among all models, attaining 52.4 mIoU. This result surpasses SegFormer-B5 by 0.9 and Swin-L-UPerNet by 1.9 mIoU, with a comparable parameter count to SegFormer-B5. The lighter FMA2D-B-SegFormer variant exhibits a similar performance trend. These results confirm that FMA’s hierarchical multi-resolution design benefits dense pixelwise tasks such as segmentation.

\subsection{Hyperparameter Study}\label{ablation}
To understand the impact of FMA's key hyperparameters, the base cell size $r$ and the coarse level rank $p$, we conducted a sensitivity analysis on the enwik-8 dataset. The model configuration remains consistent with our previous experiments, using a 6-layer Transformer with an embedding dimension of 768 and 12 attention heads.

The results in Table 6 show a clear trade-off between performance and efficiency. Increasing both $r$ and $p$ improves the model's accuracy (lower bpc) at the cost of a larger memory footprint. This behavior aligns with the theoretical memory complexity of $\mathcal{O}(r n+n p \log (n / r))$ (loglinear variant), which grows with both hyperparameters. These findings validate our choice of $p=4$ in previous experiments, as it offers a compelling balance between modeling capacity and computational cost.

\begin{table}[htbp]
\footnotesize
\centering
\begin{tabular}{lccc} 
 \toprule
    & r=16 (6 levels)& r=32 (5 levels)& r=64 (4 levels)\\
 \midrule
 p=1 & 1.321 (18.1 GB) & 1.279 (19.1 GB) & 1.270 (20.5 GB) \\
 p=2 & 1.320 (18.4 GB)& 1.272 (19.3 GB) & 1.258 (20.9 GB) \\
 p=4 & 1.307 (19.1 GB)& 1.266 (19.7 GB) & 1.256 (21.6 GB) \\
 p=8 & 1.276 (20.2 GB)& 1.262 (21.1 GB) & 1.251 (23.2 GB) \\
 \bottomrule
\end{tabular}
\caption{Autoregressive language modeling results with different $r$ and $p$ values on enwik-8 with context size $n=1024$.}
\label{mp}
\end{table}

\subsection{Efficiency Comparison} \label{efficiency}
\begin{figure*}[!ht]
    \centering
\includegraphics[width=1\textwidth]{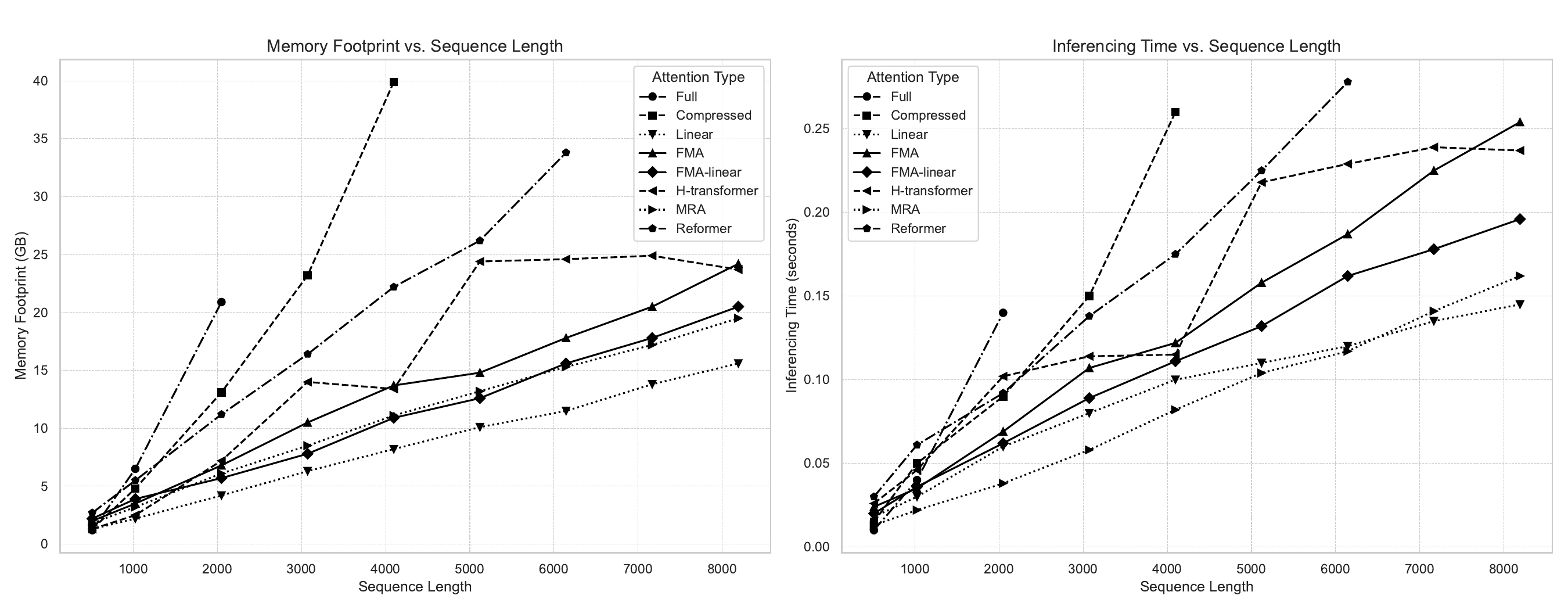}
  \caption{Efficiency comparison of attention variants on different input sizes. (left) Memory footprint. (right) Inferencing time.}
      \label{fig:efficiency}
\end{figure*}

To empirically validate the scalability of FMA, we benchmark its inference time and peak GPU memory footprint against several other attention mechanisms. We compare both FMA ($\mathcal{O} (n\log n)$) and FMA-linear ($\mathcal{O} (n)$) to quadratic baselines (Full and Compressed attention) and other popular efficient methods (Reformer, Linear, H-Transformer, and MRA). The input sequence length varies from 256 to 8192 tokens, with a consistent batch size of 64 and an embedding dimension of 768.

The left panel of Figure \ref{fig:efficiency} clearly shows the practical limitations of quadratic attention. The memory required for Full and Compressed attention grows quadratically, becoming prohibitively expensive and exceeding 40 GB at a sequence length of just 4096. In contrast, the FMA variants demonstrate far superior memory scaling. FMA-linear is among the most memory efficient methods tested, maintaining a low footprint that scales linearly and is comparable to MRA and H-transformer across all sequence lengths. The loglinear FMA also shows excellent memory growth, confirming its efficiency for long-context applications.

The inference latency results in the right panel of Figure \ref{fig:efficiency} mirror the memory analysis. As sequence length increases, the inference time for Full attention grows quadratically, making it impractical for real-time applications with long inputs. The FMA family, however, scales gracefully. FMA-linear is one of the fastest attention mechanisms, with its inference time growing slowly and linearly, remaining competitive with other highly optimized methods like Reformer and H-Transformer even at 8192 tokens. In summary, these results confirm that the FMA framework successfully translates its theoretical complexity benefits into practical performance gains. Both variants offer a robust and scalable alternative to standard self-attention, with FMA-linear among the best solutions for minimizing both memory and latency. It is also important to note that our current CUDA kernels may not have been fully optimized yet, suggesting that further performance improvements are possible.

\section{Discussion}\label{sec:discussion}

We have introduced Fast Multipole Attention, a principled,
physics-inspired replacement for quadratic self-attention that offers
(i) a full global receptive field,  
(ii) $\mathcal{O}(n\log n)$ or $\mathcal{O}(n)$ complexity, and
(iii)~a unified formulation for sequences and images.
By learning multiresolution aggregation kernels, FMA preserves fine
detail in a fixed $3r$ (1D) or $3r\times3r$ (2D) neighborhood while
capturing long-range context through learned low-rank group summaries.
Extensive experiments confirm that the model substantially improves accuracy among efficient attention mechanisms in language
(enwik8, WikiText-103) and vision
(ImageNet-1K, ADE20K) versus strong baselines such as Reformer, Swin, and SegFormer, often with substantially improved memory use.

Because far-field blocks are never materialized, both memory and compute grow
linearly with input area, making $512^{2}$ images or sequences with
$10^{5}$ tokens feasible even with limited computational resources.
The same kernel design can extend naturally to 3D inputs (e.g. videos
or volumetric medical scans) by replacing quadtrees with octrees,
suggesting a path toward efficient spatiotemporal transformers.

FMA introduces only two hyperparameters, the base cell size $r$ and the
coarse level rank $p$.
Although their default values ($r=64$ for long inputs, $p=4$)
work well across language tasks, very small inputs or highly structured data may
benefit from task-specific tuning.

Three future directions appear particularly promising:
(i) fusing FMA with rotary or relative positional encodings to improve
long-range extrapolation in language models;
(ii) incorporating sparsity aware training tricks
(low-rank adapters, progressive pruning) to further reduce FLOPs; and
(iii) exploring hybrid architectures that combine FMA’s global context
with convolutional inductive biases for low-level vision tasks
(e.g. superresolution or denoising).

\appendix

\newpage

\newpage
\bibliographystyle{abbrv} 
\bibliography{ref}
\end{document}